\documentclass[runningheads]{llncs}
\usepackage{esvect}
\usepackage[T1]{fontenc}
\usepackage{graphicx}
\usepackage{booktabs}
\usepackage[misc]{ifsym}
\newcommand{\corr}{(\Letter)}
\usepackage{amsmath}
\usepackage{hyperref}
\usepackage{caption}
\usepackage{amssymb}
\usepackage{subcaption}
\usepackage{float}
\usepackage{xcolor}
\usepackage{array}
\usepackage{csquotes}
\usepackage{adjustbox}
\usepackage{multirow}
\usepackage{xurl}
\begin{document}

\title{Train, Test, Re-evaluate: Schedule-Sensitive Evaluation of Generative Data for Hand Detection}
\titlerunning{Evaluation of Generative Data}

\author{Atmika Bhardwaj\inst{1} \corr \orcidID{0000-0002-0721-5456} \and Silvia Vock\inst{1} \and Nico Steckhan\inst{1}  \orcidID{0000-0003-0245-2046}}
\authorrunning{A. Bhardwaj et al.}
\institute{1 Federal Institute for Occupational Safety and Health (BAuA), Germany \email{\{bhardwaj.atmika, vock.silvia, steckhan.nico\}@baua.bund.de}}


\maketitle

\begin{abstract}

Generated (or synthetic) image data is increasingly used to augment or replace real training datasets when target imagery is scarce, expensive, or biased.
For hand detection, particularly in occupational safety settings, public datasets mostly contain bare hands.
This under-represents the variation in hand appearance introduced by gloves, tattoos, jewelry, and other personal protective equipment, creating a distribution shift that safety-critical applications encounter at deployment.
We test whether \emph{generative inpainting}, editing only the hand region of a real photograph to introduce accessories, can close this shift gap and improve detection of real hands at deployment.
On a paired dataset of real images and their synthetic counterparts, we evaluate YOLOv8n hand detectors across six experiments (A-F), four of which involve training (A, C, D, E) under three random seeds each, evaluate them on a real test set and on a real-gloves-only test split, and report the mean average precision (mAP) at two overlap thresholds (mAP@0.5 and mAP@0.5:0.95) along with paired statistical tests.
A two-stage experiment: train on real $\cup$ synthetic data, then fine-tune the resulting weights on real-only at a lower learning rate, directionally improves mAP@0.5 compared to the real-only baseline model on the standard real test set, and narrows the real-gloves out-of-distribution gap.
Another three-stage experiment preserves box-tightness best, achieving the highest mAP@0.5:0.95 among experiments in the study.
The synthetic-data utility for safety-critical hand detection depends on the \emph{training procedure}, and simple multi-stage experiments extract substantial real-deployment benefit from inpainted accessory data.

\keywords{Generative inpainting  \and Hand detection \and YOLOv8n \and Data augmentation \and Occupational safety monitoring \and Paired-image evaluation}
\end{abstract}

\section{Introduction}\label{sec:Introduction}

Detecting human hands in images is a foundational building block in human-machine interaction, gesture analysis, industrial (occupational) safety monitoring, and surgical workflow analysis.
Among these, occupational safety in particular requires robust detection of hands wearing accessories (gloves, band-aids, tattoos, jewelry, etc.).
The appearance shift is often under-represented in widely used public datasets, which are dominated by bare-handed gestural imagery~\cite{afifi2018,chen2025,si2026}.
The resulting mismatch between training data and deployment data, \emph{distribution shift}~\cite{kulinski2023}, is a major practical obstacle to using off-the-shelf hand detectors in safety-critical contexts~\cite{jalayer2026,sharma2025}.

Generating synthetic hand images with accessories is a promising way to close that gap.
Whether it actually works depends on two separate questions: (a)~do the synthetic images \emph{look} real, and (b)~do they reduce the synthetic-to-real gap for hands near the machine and \emph{improve} safety performance on real deployment images?
Recent work points in the same direction: whole-image realism metrics are often too coarse for object-centric tasks, hands still have a persistent simulation-to-real gap~\cite{zhao2025}, and synthetic data helps most when it is filtered, targeted, and mixed with real data, not used blindly or as a replacement.
\emph{Generative inpainting} enables selectively editing the hand of a real photograph with a generative model such that the result is the same photograph but with a gloved or accessorized hand~\cite{yang2024}.
This means that the background of the hand remains pixel-identical, producing tightly matched image pairs.
The open scientific question is whether the synthetic images produced by this method are \emph{useful} for training a real-world detector.

The deployment scenario is camera-based hazard monitoring at a dimension saw: a fixed camera observes the operator's hands as they feed and position workpieces, and the safety function requires knowing when a hand enters the danger zone so that some response can be triggered.
The goal of this work is therefore explicitly to improve detection of \emph{real} hands at deployment; synthetic imagery is a means to that end.
For every pair of synthetically generated and real test images, we compute five paired image-quality metrics to substantiate that: (a) the synthetic images are recognizable hand imagery, and (b) they differ from the originals in systematic, measurable ways.
Then, we design six YOLOv8n~\cite{varghese2024} detection experiments: four training regimes, one robustness probe without retraining, and one cross-experiment analysis.
The experiments target both the training-data composition (real-only, real$+$synthetic, synthetic-only, real-gloves-only) and the training procedure (single-stage, two-stage augment-then-fine-tune, and three-stage curriculum with progressively decreasing learning rate).
Three random seeds per experiment enable their statistical analysis.

The remainder of the paper is organized as follows.
Section~\ref{sec:methods} describes the dataset with real images~\ref{sec:dataset_real}, labeling format and the hand-inpainted dataset~\ref{sec:dataset_gen}.
The paired image comparison metrics with definitional thresholds are discussed in Section~\ref{sec:img_comparison_metrics}.
Section~\ref{sec:experiments} defines the detection experiments (A-F).
Section~\ref{sec:results} reports the results and discussion: per-experiment test mAP, statistical tests, and the cross-experiment gap analysis.
Section~\ref{sec:conclusion} discusses the broader implications and outlines directions for closing the remaining out-of-distribution gap.

\vspace{-5pt}
\section{Related Work}

\textbf{Synthetic Data for Object Detection.} The use of synthetically generated or augmented data to improve object detection models has become an active area of research, particularly in data-scarce domains.
Early approaches relied on 3D rendering engines and domain randomization to produce annotated training images, as surveyed comprehensively by Westerski et al.~\cite{Westerski2024} on synthetic data for object detection with domain randomization techniques.
More recently, diffusion-based generative models have emerged as a powerful alternative.
Li et al.~\cite{li2024} and Wołk et al.~\cite{wolk2026} demonstrated that augmenting training sets with Stable Diffusion-generated background variations significantly improved robustness for object detection.
These studies also highlight a critical open problem: how to systematically assess the quality and downstream utility of the generated images, particularly when the inpainted content must satisfy domain-specific constraints such as anatomical plausibility.

\textbf{Hand Detection.}
In the broader context of industrial safety, YOLO-based systems have been deployed for safety equipment detection, as noted by Islam et al. ~\cite{islam2024deeplearningapproachdetect}, achieving $87.7\%$ mAP for detecting helmets, goggles, jackets, and gloves using YOLOv7.
Hand detection in industrial environments is a safety-critical task with specific challenges.
A comprehensive review by Jalayer et al.~\cite{jalayer2026} surveys deep learning approaches for hand detection, segmentation, and gesture recognition in human-robot interaction, noting that YOLO-based architectures consistently outperform alternatives in balancing accuracy and speed.
In the work by Hubert et al.~\cite{hubert2025}, a YOLOv8n-based model was employed to detect the hands of operators interacting with an industrial cobot across multiple camera viewpoints.

\vspace{-8pt}
\section{Methods}\label{sec:methods}
\vspace{-5pt}
\subsection{Dataset: Real Hand Images}\label{sec:dataset_real}

The images were recorded by the Fraunhofer Institute for Manufacturing Engineering and Automation (IPA), Germany, at a single instrumented dimension saw workstation with a fixed camera viewpoint, and the train/validation/test split was already partitioned by them.
Hand appearance is bare or gloved; the real dataset contains no jewelry and no tattoos.
This matches the occupational-safety motivation, since gloves are the mandated protective equipment at workstations and hide the skin-texture cues (creases, knuckles, fingernails) that hand detectors rely on.
This dataset comprises $6,507$ training, $2,085$ validation, and $3,591$ test RGB images (\texttt{.PNG} format), each with a \texttt{.txt} file annotating a single \texttt{hand} class in YOLO format.
Note that the validation set only has bare hands.
Multi-hand images contribute one line per hand in their respective \texttt{.txt} files.

\vspace{-5pt}
\subsection{Generated Dataset: Inpainted Hand Images}\label{sec:dataset_gen}

The generated dataset is produced with a recent tool developed in our group \emph{SemProbe}\footnote{\url{https://github.com/steckhan/semrob}}~\cite{steckhan2026semantic}, an open-source tool for semantic robustness probing of safety-critical object detectors via controlled diffusion-based inpainting.
SemProbe modifies a localized region of a real image while leaving the surrounding scene as pixel-identical as possible.
Whether this localization assumption holds in practice is itself a question that this paper addresses.

For each real image, the inpainting pipeline proceeds in three stages.
\emph{First}, the hand region is segmented automatically using GroundingDINO~\cite{liu2024} conditioned on the text prompt \texttt{"hand"}, followed by SAM2~\cite{ravi2024} to refine the bounding box proposal into a pixel-level binary mask.
The mask is then dilated by a small margin to ensure that the inpainting model has sufficient context at the hand boundary; manual correction is applied where the automatic mask fails (e.g., on heavily occluded hands).
\emph{Second}, the masked region is regenerated using FLUX.2~[klein], a 9\, B-parameter rectified-flow transformer for mask-based image editing, run on NVIDIA RTX 6000 Ada at $1024\times1024$ resolution to preserve data sovereignty.
Generation is conditioned on prompts (such as \enquote{put jewelry only on the given hands and fingers} or \enquote{put a colorful tattoo of any color only on the given hand}), and generation parameters (seed, steps, CFG scale, denoise strength, sampler) are set to their default values across the dataset~\cite{steckhan2026semantic}.
These prompts act as appearance perturbations that diversify the training data distribution.

\begin{figure}[!t]
  \vspace{-10pt}
  \centering
  \includegraphics[width=0.85\textwidth]{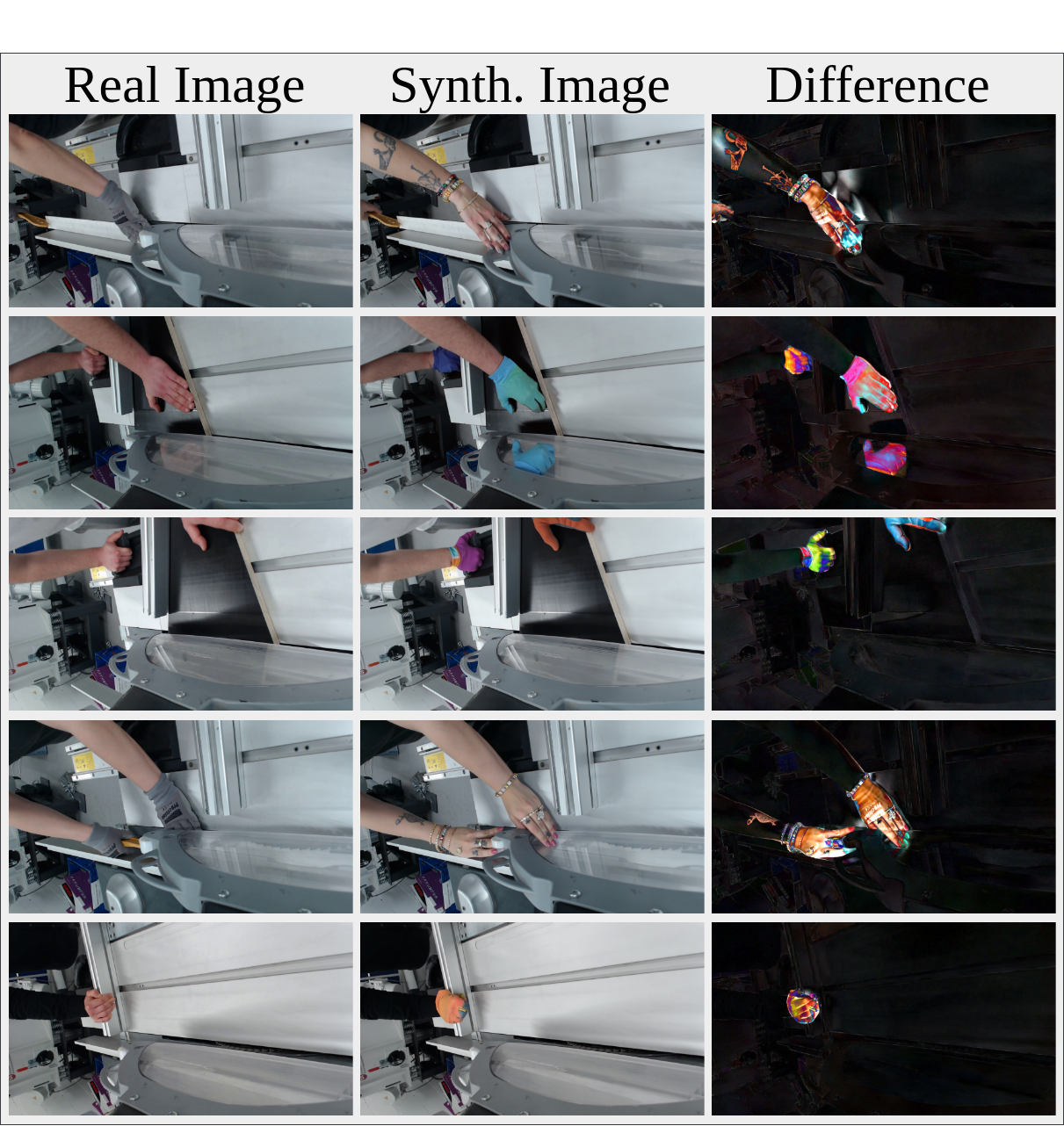}
  \caption{Five randomly selected image pairs (one per row).
  Left: real.
  Centre: its synthetic counterpart (same background, hands inpainted with accessories).
  Right: amplified pixel-wise absolute difference, computed as $D = \mathrm{clip}\bigl(3 \cdot |I_{\text{real}} - I_{\text{synth}}|,\,0,\,255\bigr)$ visualized with the ``hot'' colormap; bright regions mark pixels that the inpainting modified.}
  \vspace{-20pt}
  \label{fig:side_by_side}
\end{figure}

The pipeline produces a synthetic training set of $6,507$ images and a synthetic test set of $3,591$ images, paired one-to-one with the corresponding real splits.
Fig.~\ref{fig:side_by_side} shows five of these real images, their synthetically generated image, and the pixel-wise difference in those images.
Bounding box labels for the synthetic images are inherited directly from their real counterparts, since the inpainted hand occupies the same image region as the original ones by construction.
This (a) enables evaluation of both overall per-paired-image metrics as well as hand-region-restricted metrics reported in Section~\ref{sec:img_comparison_metrics} and (b) preserves YOLO-format compatibility across all detection experiments (A-F) in Section~\ref{sec:experiments}.

\vspace{-5pt}
\subsection{Image Comparison Metrics}\label{sec:img_comparison_metrics}
\begin{figure}[htbp]
  \vspace{-10pt}
  \centering
  \includegraphics[width=0.9\textwidth]{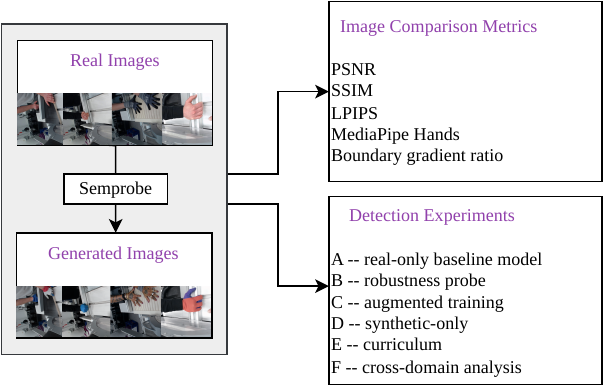}
  \caption{An illustrative workflow of this work.}
  \vspace{-10pt}
  \label{fig:workflow}
\end{figure}

A systematic workflow is shown in Fig.~\ref{fig:workflow}
We evaluate five paired-image quality metrics computed on the test corpus ($3,591$ paired real and synthetic images) to quantify \emph{how much} of each image was edited and \emph{where}.
\begin{enumerate}
  \item \textbf{PSNR (Peak Signal-to-Noise Ratio).} Measures raw pixel-level difference~\cite{faragallah2021}.
  It computes the average squared error between every pixel in the two images and converts it to a decibel scale.
  Values above 30~dB typically indicate differences are barely visible to humans.
  \item \textbf{SSIM (Structural Similarity Index).} Slides a small window across both images and compares local brightness, contrast, and texture~\cite{rouse2008}.
  Range $[-1, 1]$; values close to 1 indicate a nearly identical structure.
  \item \textbf{LPIPS (Learned Perceptual Image Patch Similarity).} Passes both images through a pretrained \texttt{VGG-16} neural network, extracts features at multiple internal layers, and computes how different those features are~\cite{zhang2018}.
  This captures \enquote{does it look different to a human} rather than \enquote{are the pixels different}.
  Range $[-1, 1]$; 0 = identical images.
  \item \textbf{Anatomical Metrics (MediaPipe Hands).} Google’s MediaPipe Hand Landmarker model, which is third-party pretrained on a large bare-handed dataset (or palms), is run on every real and synthetic test image~\cite{zhang2020}.
  It is configured to allow up to ten hands per image and to reject palm detections below a confidence of $0.5$.
  We record (i) whether at least one hand is detected and its bounding box, (ii) its detection confidence score, and (iii) the 21 hand landmark coordinates (wrist, four points per finger).
  Note that accessorized hands (gloves, jewelry, and tattoos) partially cover, hide, or distort the visual features (creases, knuckles, fingernails) that the model relies on to localize key points.
  \item \textbf{Boundary gradient ratio.} Inpainting pipelines that edit only a region of an image can leave a visible seam at the boundary between the unchanged background and the freshly synthesized content.
  To detect such seams, we extract the $15$-pixel-wide strip $S_B$ straddling each labeled bounding box edge (half inside, half outside), compute the mean Sobel gradient inside that strip on the real image and on the synthetic image~\cite{sobel2014}, and take their ratio (Eq.~\ref{eq:boundary-ratio}).
  For each labeled hand bounding box $B$, let $\|\nabla I(u,v)\|$ denote the Sobel gradient magnitude at pixel $(u,v)$ of a gray-scale image $I$.
  The mean strip gradients on the real and synthetic versions of the same image are
  \begin{equation}
  \vspace{-3pt}
  \begin{split}
    \bar{g}_{\text{real}}(B)  & = \frac{1}{|S_B|}\!\!\sum_{(u,v) \in S_B}\!\!
      \|\nabla I_{\text{real}}(u,v)\|~~, \\ 
    \bar{g}_{\text{synth}}(B) & = \frac{1}{|S_B|}\!\!\sum_{(u,v) \in S_B}\!\!
      \|\nabla I_{\text{synth}}(u,v)\|~~,
    \end{split}
    \end{equation}
    and the per-box boundary gradient ratio is
    \begin{equation}
    r(B) = \frac{\bar{g}_{\text{synth}}(B)}{\bar{g}_{\text{real}}(B)}~~.
    \label{eq:boundary-ratio}
    \end{equation}
    A value of $r(B) = 1.0$ means the synthetic strip has the same average edge content as the real strip at the same location (invisible seam); $r(B) > 1$ means the inpainting introduced stronger edges along the boundary; $r(B) < 1$ means it smoothed over edges that were present in the real image.
    We observe a value of $1.088 \pm 0.225$, which sits just above the invisible-seam baseline, with a long upper tail.
    Meaning, hand pairs show only a mild seam, but a minority of them have visibly stronger boundary gradients.
\end{enumerate}

\begin{figure}[htbp]
\vspace{-30pt}
  \includegraphics[width=\textwidth]{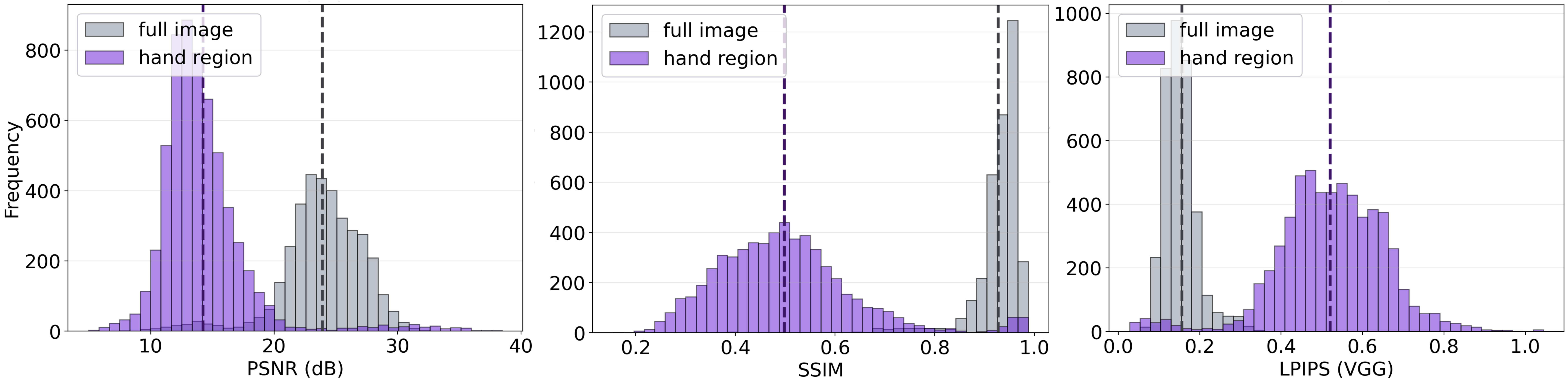}
  \caption{Per-image PSNR, SSIM, and LPIPS for full image (gray; $3,591$ pairs) vs hand region crops (purple; $5{,}884$ pairs).
  The hand region shows more pairs than full images because many frames have $2$ hands.
  Dashed vertical lines mark their mean values.}
  \label{fig:metrics}
  \vspace{-20pt}
\end{figure}

We compute PSNR, SSIM, and LPIPS metrics twice: once on the entire image to test whether the edit stayed local and once on the hand region to quantify the magnitude of the intended change, as shown in Table~\ref{tab:diag-perception} and in Fig.~\ref{fig:metrics}.
A hand region is a hand crop defined by the YOLO bounding boxes in the label (\texttt{.txt}) files, which are rectangles bounded by the ground-truth bounding boxes.
It isolates the inpainted region from the untouched background and provides a more precise evaluation of inpainting quality.
The decrease in values from the full image to the hand region for every metric reflects the fact that the generator modifies only the hand region; the unmodified background dominates the full-image scores.

\begin{table}[H]
\centering
\footnotesize
\vspace{-15pt}
\caption{PSNR (Peak Signal-to-Noise Ratio), SSIM (Structural Similarity Index) and LPIPS (Learned Perceptual Image Patch Similarity) between each real image and its synthetic counterpart.
\emph{Min/Max} are the theoretical range of the metric.
\emph{Identical} is the value the metric returns when the two images are byte-identical (or perceptually indistinguishable).
\emph{Full image/Hand region} is mean $\pm$ standard deviation across $3{,}591$ paired test images.
}
\setlength{\tabcolsep}{5.5pt}
\begin{tabular}{llllll}
\toprule
Metric      & Min    & Max       & ``Identical''             & Full image        & Hand region \\
\midrule
PSNR (dB)   & $0$    & $\infty$  & $\ge 40$ (near-lossless)  & $23.91 \pm 3.31$  & $14.23 \pm 3.73$\\
SSIM        & $-1$   & $1$       & $1.0$                     & $0.927 \pm 0.055$ & $0.498 \pm 0.138$ \\
LPIPS (VGG) & $0$    & $\sim\!1$ & $0$                       & $0.156 \pm 0.060$ & $0.521 \pm 0.136$ \\
\bottomrule
\end{tabular}
\label{tab:diag-perception}
\end{table}

Table~\ref{tab:mediapipe} reports anatomical plausibility.
The \emph{detection rate} ($0.414$ real vs $0.314$ synthetic) is the fraction of images in which MediaPipe finds at least one hand.
Both numbers are far below $100\%$, which means MediaPipe fails to detect many real (gloved) hands, too.
These drops are partly because the generator may produce hands that the palm detector simply cannot recognize as palm-shaped.
\emph{Average confidence} ($0.965$ vs $0.932$) is the mean of the palm detector's per-hand confidence score, and \emph{anatomical-pass rate} ($0.947$ vs $0.997$) is a loose sanity check that all 21 landmarks lie inside the image and that each finger has non-zero length.
These are computed only over the hands that MediaPipe found (not across all images).
The \emph{landmark deviation} ($0.222 \pm 0.187$) is the only per-real-synthetic-pair metric.
For each test image where MediaPipe detected at least one hand on both the real and synthetic sets, we match those hands via the Hungarian algorithm implementation in SciPy on hand-center distance~\cite{kuhn1955}.
For a matched pair, 21 landmark coordinates are written as $\{p_i^{\text{real}}\}_{i=1}^{21}$ and $\{p_i^{\text{synth}}\}_{i=1}^{21}$, with $p_i = (x_i, y_i)$ in image-pixel coordinates.
The per-pair mean Euclidean landmark shift ($\bar{l}$) and the real hand's landmark bounding box diagonal ($\mathrm{diag}$) are
\begin{equation}
\bar{l}
   = \frac{1}{21}\sum_{i=1}^{21}
     \bigl\|p_i^{\text{real}} - p_i^{\text{synth}}\bigr\|_2~~,
\end{equation}
and
\begin{equation}
\mathrm{diag}
   = \sqrt{\bigl(\max_i x_i^{\text{real}} - \min_i x_i^{\text{real}}\bigr)^{\!2}
         + \bigl(\max_i y_i^{\text{real}} - \min_i y_i^{\text{real}}\bigr)^{\!2}}~~,
\end{equation}
and the normalized, scale-invariant landmark deviation for the pair is $\delta = \bar{l}/\mathrm{diag}$.

The value reported in Table~\ref{tab:mediapipe} is the mean of $\delta$ across all matched hand pairs in the test corpus.
Values $\delta < 0.05$ are conventionally taken as \enquote{preserved pose}; our observed mean of $0.222$ is more than $4\times$ that threshold.
This number, therefore, conflates two effects we cannot fully separate from the aggregate number: (i) genuine pose drift introduced by the inpainting, and (ii) reduced MediaPipe reliability on accessory-bearing synthetic hands.
The number should be read as an upper bound on (i), not a clean measurement of it.
For instance, for image pairs where the real hand is bare and the synthetic hand is gloved, MediaPipe localizes landmarks accurately on the real side but \emph{extrapolates} them from the hand silhouette on the synthetic side; even if no real pose change has occurred, the metric would still report non-zero deviation simply because MediaPipe is reading the two images with different reliability.
The $\pm 0.187$ standard deviation is large, indicating substantial spread; many pairs sit near zero drift, while others are well above the mean.
A future analysis restricted to bare-hand pairs (excluding the gloved range) would allow us to separate genuine pose drift from MediaPipe-reliability noise.
With these diagnostics established, the detection experiments in the next Section~\ref{sec:results} evaluate whether these differences matter for training hand detectors.

\begin{table}[H]
\vspace{-20pt}
\centering
\footnotesize
\caption{Anatomical plausibility via MediaPipe Hands for both real and synthetic images.
}
\setlength{\tabcolsep}{3.2pt}
\begin{tabular}{lllllll}
\toprule
MediaPipe Metric                         & Region           & Min & Max       & ``Identical''   & Real & Synthetic\\
\midrule
Detection rate       & whole image set  & $0$ & $1.0$     & ---             & $0.414$    & $0.314$ \\
Avg.\ confidence     & detected hands   & $0$ & $1.0$     & ---             & $0.965$    & $0.932$ \\
Anatomical-pass rate & detected hands   & $0$ & $1.0$     & $1.0$           & $0.947$    & $0.997$ \\
Landmark deviation   & detected hands   & $0$ & $\sim\!1$ & $0$ (same pose) & \multicolumn{2}{l}{$0.222 \pm 0.187$} \\
\bottomrule
\end{tabular}
\label{tab:mediapipe}
\end{table}

\vspace{-22pt}
\section{Detection Experiments}\label{sec:experiments}
\vspace{-2pt}
\subsection{Experimental Setup (A-F)}

We report six experiments; each tests a single hypothesis.
Four involve training (A, C, D, E); Exp~B reuses the weights from Exp~A without retraining, and Exp~F aggregates the results of A--E.
Exp~C and Exp~E are multi-stage experiments in sequence.
The second and subsequent stages do not start from fresh weights but \emph{warm-start}, i.e., they initialize from the best checkpoint saved by the preceding stage.
So each stage refines what the previous one learned instead of relearning from scratch.
All training experiments use three random seeds ($\{42, 43, 44\}$) and are reported as the mean $\pm$ standard deviation across seeds.
The list of experiments is as follows:
\vspace{-5pt}
\begin{itemize}
  \item \textbf{Exp~A -- Baseline (Real $\rightarrow$ Real).} Train on real, test on real.
  Provides the upper-bound reference.
  \item \textbf{Exp~B -- Robustness probe.} No retraining; evaluate the three Exp~A models on two test splits: synthetic data and real gloves to measure out-of-distribution (OOD) behavior.
  \item \textbf{Exp~C -- Augmented training (two-stage).} \emph{Stage 1:} train on real $\cup$ synthetic for up to $225$ epochs at a learning rate of $\mathrm{lr}_0=10^{-2}$.
  \emph{Stage 2:} fine-tune Stage 1's training weights on real-only data for $20$ epochs at a smaller learning rate of $\mathrm{lr}_0=10^{-3}$.
  Tests whether a two-stage augment-then-fine-tune procedure delivers usable accuracy improvements over the real-only baseline.
  \item \textbf{Exp~D -- Synthetic-only training.} Train on synthetic data only for $75$ epochs; tested on real.
  It is the strongest test of synthetic-data utility: can a model that has never seen a real image detect real hands?
  \item \textbf{Exp~E -- Curriculum (Real $\to$ Real$+$Synth $\to$ Real gloves).} Three-stage training.
  \emph{Stage 1:} real-only, $75$ epochs, $\mathrm{lr}_0=10^{-2}$.
  \emph{Stage 2:} warm-start from Stage 1, train on real $\cup$ synthetic at $\mathrm{lr}_0=10^{-3}$.
  \emph{Stage 3:} warm-start from Stage 2, fine-tune on the real-gloved-only training images at $\mathrm{lr}_0=10^{-4}$.
  This tests whether an explicit curriculum preserves box tightness on the real test set while closing the real-gloves OOD gap.
  \item \textbf{Exp~F -- Cross-domain meta-analysis.} No training; aggregates the results of trained models into signed gap metrics (real-to-OOD on synthetic gloves, real-to-OOD on real gloves, synthetic-to-real, Exp~C augmentation benefit on real test and real gloves, Exp~E curriculum benefit on real test and real gloves).
\end{itemize}

\begin{table}[H]
\centering
\footnotesize
\vspace{-30pt}
\caption{Per-experiment hyperparameters.
}
\setlength{\tabcolsep}{10pt}
\begin{tabular}{lllll}
\toprule
Experiment    & Init.\ weights         & Epochs  & $\mathrm{lr}_0$ & \# stages \\
\midrule
A             & \texttt{yolov8n.pt}    & 75  & $10^{-2}$ & $1$ \\
B             & Exp~A                  & --  & --        & $0$ (eval only) \\
C: Stage~1    & \texttt{yolov8n.pt}    & 225 & $10^{-2}$ & $2$ \\
C: Stage~2    & Stage~1                & 20  & $10^{-3}$ &   \\
D             & \texttt{yolov8n.pt}    & 75  & $10^{-2}$ & $1$ \\
E: Stage~1    & \texttt{yolov8n.pt}    & 75  & $10^{-2}$ & $3$ \\
E: Stage~2    & Stage~1                & 75  & $10^{-3}$ &   \\
E: Stage~3    & Stage~2                & 30  & $10^{-4}$ &   \\
\bottomrule
\end{tabular}
\vspace{-15pt}
\label{tab:hyperparams}
\end{table}

\noindent Beyond warm-starting, each later stage typically uses a different training set, a smaller learning rate, or both, so that it \emph{refines} rather than \emph{overwrites} the previous stage.
The model that gets evaluated is the final stage's best checkpoint.
Table~\ref{tab:hyperparams} shows the per-experiment hyperparameters.
The epochs specify only an upper bound on the schedules rather than a fixed stopping point; the weights corresponding to the best epoch on real validation data are saved.
All training experiments have batch size $16$, image size $640$, AdamW optimizer, three random seeds, the standard YOLOv8n augmentation defaults, and ReduceLROnPlateau (i.e., if the validation loss fails to improve after a fixed patience of $15$ consecutive epochs, $\mathrm{lr_0}$ is multiplied by $0.5$).
It acts as a safeguard against noisy real-only validation data when synthetic samples are in the training set.
Table~\ref{tab:exp-data} summarizes which data each experiment trains and tests on, including the number of images per split.
Hands featuring various accessories appear only in the synthetic images: the real dataset includes bare and gloved hands.
The validation split is the same set of $2{,}085$ real images for all.

\begin{table}[H]
\centering
\footnotesize
\vspace{-20pt}
\caption{Data flow per experiment.
\enquote{Stage 1}/\enquote{Stage 2}/\enquote{Stage 3} refer to training stages within the multi-stage Exps~C and E.
$N$ refers to the number of images.
The $1{,}000$ images real-gloves split is a subset of the standard real test set.}
\adjustbox{width=1.0\textwidth}{
\begin{tabular}{lllll}
\toprule
Experiment         & Train data ($N$)               & Validation ($N$)  & Test data ($N$)\\
\midrule
A                  & real ($6{,}507$)               & real ($2{,}085$)  & real ($3{,}591$)  \\
B                  & --                             & --                & synthetic ($3{,}591$) + real gloves ($1{,}000$) \\
C: Stage 1 & real $\cup$ synth ($13{,}014$) & real ($2{,}085$)  & --   \\
C: Stage 2 & real ($6{,}507$)               & real ($2{,}085$)  & real ($3{,}591$) + real gloves ($1{,}000$) \\
D          & synth ($6{,}507$)              & real ($2{,}085$)  & real ($3{,}591$)   \\
E: Stage 1 & real ($6{,}507$)               & real ($2{,}085$)  & --   \\
E: Stage 2 & real $\cup$ synth ($13{,}014$) & real ($2{,}085$)  & --    \\
E: Stage 3 & real gloves ($911$)            & real ($2{,}085$)  & real ($3{,}591$) + real gloves ($1{,}000$) \\
\bottomrule
\end{tabular}
}
\label{tab:exp-data}
\end{table}

\vspace{-25pt}
\section{Results and Discussion}\label{sec:results}
\vspace{-1pt}
\subsection{Test Results}

\begin{figure}[!htbp]
\centering
  \includegraphics[width=\textwidth]{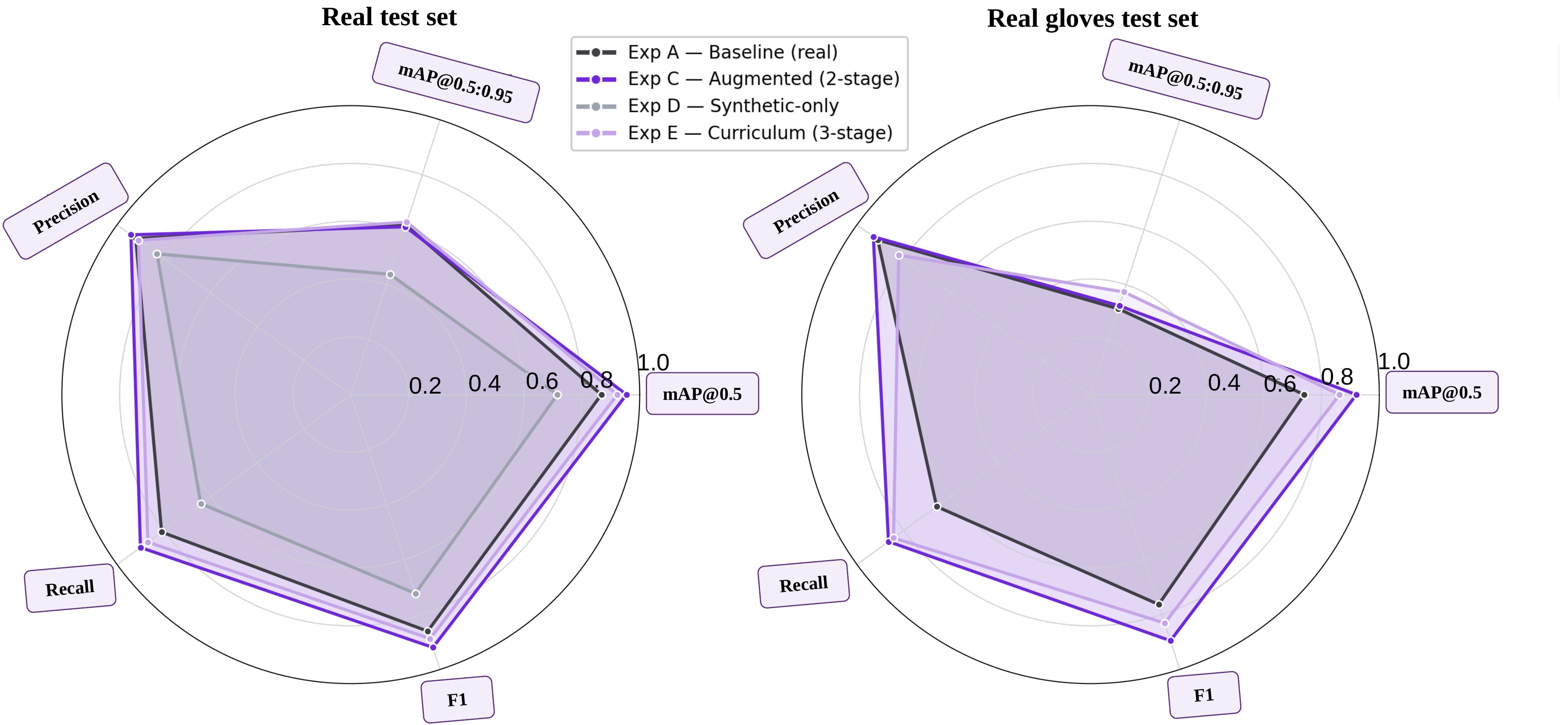}
  \caption{Five-metric radar comparison of the trained detectors on (left)~the real test set with $3,591$ and (right)~its real-gloves subset with $1{,}000$ images.
  All metrics are shown with radial axis locked to $[0, 1]$; larger polygon area indicates better all-around performance; values are taken from Table~\ref{tab:cross-exp}.
  Since trained Exp~A was used to test on real gloves only in Exp~B, those values of Exp~B are shown.
  }
  \label{fig:radar_plot}
  \vspace{-15pt}
\end{figure}

For each trained detector and each test split, we report:
(a) \emph{mAP@0.5} (mean Average Precision at intersection over union (IoU) threshold $0.5$, i.e., a prediction counts as correct if it overlaps a ground-truth box by at least $50\%$),
(b) \emph{mAP@0.5:0.95} (mAP averaged over IoU thresholds $\{0.5, 0.55, \dots, 0.95\}$, which is stricter, penalizes imprecise box edges),
(c) \emph{precision},
(d) \emph{recall} at the default confidence threshold of $0.5$,
and (e) \emph{F1} (their harmonic mean).
Table~\ref{tab:cross-exp} shows per-experiment test-set numbers: each row is a mean $\pm$ standard deviation of three seed runs.
After each epoch, the model is scored on the validation split by a scalar, $0.1\cdot\mathrm{mAP@0.5} + 0.9\cdot\mathrm{mAP@0.5{:}0.95}$; the highest scoring epoch is saved as \texttt{best.pt}.
Fig.~\ref{fig:radar_plot} illustrates these five metrics.
The two rows in Exp~B correspond to the same weights from Exp~A evaluated on the synthetic-gloves (sg) and real-gloves (rg) test sets (shown on the right in Fig.~\ref{fig:radar_plot}).
Exp~C achieves the best mAP@0.5 on both real and real-gloves splits; Exp~E achieves the best mAP@0.5:0.95 on both.
The synthetic set contains real detection signal, as synthetic-only training (Exp~D) still reaches $83\%$ ($0.715 / 0.867$) of the real-only mAP@0.5 and should be used as a partial substitute (or to supplement) real data.

\begin{table}[h]
\centering
\caption{Per-experiment test-set results with three random seeds, reported as mean $\pm$ standard deviation.
Acronyms sg and rg refer to the synthetic-gloves and real-gloves test sets, respectively.
r$^\dagger$ refers to reusing Exp~A's trained models with no retraining.
r$+$s $\to$ r denotes two-stage schedule: train on real $\cup$ synthetic, then fine-tune on real-only.
r$\to$r$+$s$\to$rg denotes three-stage schedule: real, then real $\cup$ synthetic, then real gloves.
Bold cells mark the best result in each (test split, metric) column among trained detectors.
}
\adjustbox{width=1.0\textwidth}{
\begin{tabular}{lllccccc}
\toprule
                   &                                     &      & \multicolumn{2}{c}{mAP}                             &                 &                 & \\
\cmidrule(lr){4-5}
Exp.               & Train                               & Test & @0.5                     & @0.5:0.95                & P               & R               & F1 \\
\midrule
A                  & real (r)                            & r    & $0.867\pm0.036$          & $0.617\pm0.023$          & $0.920\pm0.005$ & $0.809\pm0.033$ & $0.861\pm0.018$ \\
\midrule
\multirow{2}{*}{B} & \multirow{2}{*}{r$^\dagger$}               & sg   & $0.394\pm0.003$          & $0.205\pm0.006$          & $0.552\pm0.009$ & $0.437\pm0.009$ & $0.487\pm0.008$ \\
                   &                                     & rg   & $0.740\pm0.076$          & $0.312\pm0.041$          & $0.911\pm0.003$ & $0.659\pm0.054$ & $0.763\pm0.034$ \\
\midrule
\multirow{2}{*}{C} & \multirow{2}{*}{r$+$s $\to$ r}      & r    & $\mathbf{0.954\pm0.004}$ & $0.611\pm0.006$          & $0.941\pm0.014$ & $0.900\pm0.009$ & $0.920\pm0.010$ \\
                   &                                     & rg   & $\mathbf{0.921\pm0.011}$ & $0.323\pm0.004$          & $0.930\pm0.014$ & $0.865\pm0.024$ & $0.895\pm0.016$ \\
\midrule
D                  & synthetic (s)                       & r    & $0.715\pm0.042$          & $0.438\pm0.031$          & $0.829\pm0.044$ & $0.642\pm0.034$ & $0.723\pm0.034$ \\
\midrule
\multirow{2}{*}{E} & \multirow{2}{*}{r$\to$r$+$s$\to$rg} & r    & $0.922\pm0.014$          & $\mathbf{0.628\pm0.009}$ & $0.909\pm0.018$ & $0.868\pm0.018$ & $0.888\pm0.014$ \\
                   &                                     & rg   & $0.862\pm0.073$          & $\mathbf{0.374\pm0.042}$ & $0.821\pm0.073$ & $0.843\pm0.080$ & $0.831\pm0.069$ \\
\bottomrule
\end{tabular}
}
\label{tab:cross-exp}
\end{table}

\subsection{Statistical Significance}
\label{sec:statistical_significance}

Cross-experiment significance is established with paired $t$-tests on per-seed mAP arrays at $n=3$, with Cohen's $d$~\cite{cohen1988statistical} as the standardized effect size in Table~\ref{tab:stat-tests}.
Each detection experiment is run three times with different random seeds, and seed indices are matched between experiments (therefore, they are \emph{paired}; run~$i$ of Exp~A and run~$i$ of Exp~C share the same seed).
Any noise introduced by the random seed (initialization, data shuffling, dropout sampling) is shared between the two members of each pair and cancels out in the difference.
For a pair of experiments $X$ and $Y$, let $x_i$ and $y_i$ denote the per-seed test-set mAP of $X$ and $Y$, respectively, and let $d_i = y_i - x_i$ be the per-seed difference.
With $n = 3$ seeds, we compute the mean difference ($\bar{d}$), its sample standard deviation ($s_d$), and the resulting $t$-statistic, $t = \bar{d}/(s_d / \sqrt{n})$.
Under the null hypothesis that the two experiments have the same expected mAP, $t$ follows a Student-$t$ distribution with $n - 1 = 2$ degrees of freedom.
The $p$-value in the table is the two-sided probability of observing a $|t|$ at least as extreme as ours under that null.
\texttt{scipy.stats.ttest\_rel(y, x)} computes both $t$ and $p$ in one call and returns them, shown in Table~\ref{tab:stat-tests}.
A small $p$-value (conventionally $p < 0.05$~\cite{Giovanni2020}) is evidence that the observed difference is unlikely to be a coincidence of seed selection but indicates statistical significance.

At $n = 3$, however, the paired $t$-test has limited statistical significance.
Even genuinely large effects can fail to reach $p < 0.05$ simply because three samples leave few degrees of freedom to estimate $s_d$ precisely.
To resolve this, we also report Cohen's $d$, a standardized measure of effect size that is independent of $n$, as $\bar{d}/s_d$.
It expresses the mean difference in units of the standard deviation of the differences.
Cohen~\cite{cohen1988statistical} proposed the following magnitude bands, which we adopt: $|d| < 0.2$ negligible, $|d| \approx 0.5$ medium, $|d| \approx 0.8$ large, and $|d| > 1.2$ very large.
A row with $p > 0.05$ but $|d| > 0.8$ should be read as the effect is large in standardized units, but the samples with three seeds are not enough to rule out chance; a row with $p < 0.05$ and $|d| < 0.2$ would read as the effect is statistically detectable but practically trivial.

In Table~\ref{tab:stat-tests}, \emph{Block 1} compares the real trained detectors (from Exps~A and C) on the real test split.
Exp~C beats Exp~A directionally on mAP@0.5 ($+0.087$, large effect size $d=2.06$) and the mAP@0.5:0.95 difference between Exps~C and A is statistically indistinguishable from zero ($-0.006$).
A possible explanation is that Stage 1 gives the model broad exposure to synthetic variants; Stage 2 calibrates box edges against real-image gradients while the small step size (learning rate) preserves Stage 1's broad prior.
Exp~E beats Exp~C significantly on mAP@0.5:0.95 ($+0.017$, $d=5.29$).
Training on synthetic data alone drops mAP@0.5 and also mAP@0.5:0.95 by $-15.2$ and $-17.9$ points, respectively, below Exp~A.
Similarly, \emph{Block 2} measures the OOD closure on real-gloves: both Exps~C and E achieve substantial glove accuracy in comparison to the real-trained baseline Exp~B ($+0.181$ and $+0.122$ mAP@0.5, respectively); the two are not statistically distinguishable from each other at $n=3$ on this real-gloves split.

\vspace{-5pt}
\subsection{Cross-domain Gaps (Exp~F)}\label{sec:exp_f}

Table~\ref{tab:exp-f-gaps} reports the signed cross-domain gap metrics from the per-experiment results of Exp~F.
The same real-trained Exp~A loses $47.3$ mAP@0.5 points on the synthetic-gloves test set but only $12.7$ points on the real-gloves test set: the synthetic test set overstates the deployment OOD gap by roughly $3.7\times$.
Training on synthetic data alone (Exp~D) costs $15.2$ points on the real test set.
The two-stage augmentation (Exp~C) gains $8.7$ points on the real test set and $18.1$ points on real gloves that Exp~A would otherwise leave open: synthetic data helps most where the domain shift occurs.
Exp~E gains $5.5$ and $12.2$ points on the same comparisons but gives the tightest boxes, as its progressively-decreasing learning rate preserves the strict-IoU regression weights learned on real-only data in Stage~1.
This makes Exp~C better suited for detection coverage and Exp~E for localization quality.

\begin{table}[!t]
\centering
\footnotesize
\caption{Paired $t$-tests on per-run mAP arrays ($n=3$).
\textit{Mean diff.} is (second experiment mean) $-$ (first experiment mean).
Mathematical context of $p$ and Cohen's $d$ are discussed in Section~\ref{sec:statistical_significance}
}
\setlength{\tabcolsep}{10pt}
\begin{tabular}{llccc}
\toprule
Comparison & Metric & Mean diff. & $p$ (paired $t$) & Cohen's $d$ \\
\midrule
\multicolumn{5}{l}{\emph{Real test split (comparing trained detectors):}} \\
\multirow{2}{*}{A vs C}   & mAP@0.5      & $\mathbf{+0.087}$ & $0.071$ & $\mathbf{+2.06}$ \\
                          & mAP@0.5:0.95 & $-0.006$          & $0.766$ & $-0.20$ \\
\midrule
\multirow{2}{*}{A vs D}   & mAP@0.5      & $-0.152$          & $0.003$ & $-11.29$ \\
                          & mAP@0.5:0.95 & $-0.179$          & $0.001$ & $-15.91$ \\
\midrule
\multirow{2}{*}{A vs E}   & mAP@0.5      & $+0.055$          & $0.133$ & $+1.42$ \\
                          & mAP@0.5:0.95 & $+0.011$          & $0.620$ & $+0.34$ \\
\midrule
\multirow{2}{*}{C vs D}   & mAP@0.5      & $-0.239$          & $0.013$ & $-4.99$ \\
                          & mAP@0.5:0.95 & $-0.173$          & $0.019$ & $-4.13$ \\
\midrule
\multirow{2}{*}{C vs E}   & mAP@0.5      & $-0.032$          & $0.112$ & $-1.57$ \\
                          & mAP@0.5:0.95 & $\mathbf{+0.017}$ & $0.012$ & $\mathbf{+5.29}$ \\
\midrule
\multicolumn{5}{l}{\emph{Real-gloves test split (OOD closure):}} \\
\multirow{2}{*}{B vs C}   & mAP@0.5      & $\mathbf{+0.181}$ & $0.087$ & $\mathbf{+1.83}$  \\
                          & mAP@0.5:0.95 & $+0.011$          & $0.759$ & $+0.20$  \\
\midrule
\multirow{2}{*}{B vs E}   & mAP@0.5      & $\mathbf{+0.122}$ & $0.203$ & $\mathbf{+1.08}$  \\
                          & mAP@0.5:0.95 & $+0.061$          & $0.211$ & $+1.05$  \\
\midrule
\multirow{2}{*}{C vs E}   & mAP@0.5      & $-0.058$          & $0.425$ & $-0.57$  \\
                          & mAP@0.5:0.95 & $+0.050$          & $0.215$ & $+1.03$  \\
\bottomrule
\end{tabular}
\vspace{-15pt}
\label{tab:stat-tests}
\end{table}

\begin{table}[!t]
\centering
\small
\caption{Cross-domain gap metrics from Exp~F.
\emph{Baseline OOD gaps:} real-trained Exp~A weights compared with synthetic-gloves and real-gloves test set (Exp~B).
\emph{Cost of replacing:} compares the models trained on synthetic data alone (Exp~D) instead of real (Exp~A) data.
\emph{Exp~C benefits:} Exp~C compared with real and real-gloves test set.
\emph{Exp~E benefits:} same comparison with mAP@0.5 of Exp~E.}
\setlength{\tabcolsep}{15pt}
\begin{tabular}{lcc}
\toprule
Gap                                    & Comparison (mAP@0.5) & Difference \\
\midrule
\multicolumn{3}{l}{\emph{Baseline OOD gaps (Exp~A weights, different test splits):}} \\
Exp~A real $\to$ synth. gloves         & $0.8672 - 0.3938$    & $+0.4734$ \\
Exp~A real $\to$ real gloves           & $0.8672 - 0.7400$    & $+0.1272$ \\
\midrule
\multicolumn{3}{l}{\emph{Cost of replacing real with synthetic (real test):}} \\
synth-only $-$ real-only               & $0.7152 - 0.8672$    & $-0.1520$ \\
\midrule
\multicolumn{3}{l}{\emph{Exp~C (two-stage augmented) benefits:}} \\
Exp~C $-$ Exp~A (real test)            & $0.9542 - 0.8672$    & $+0.0870$ \\
Exp~C $-$ Exp~B (real gloves)          & $0.9206 - 0.7400$    & $+0.1806$ \\
\midrule
\multicolumn{3}{l}{\emph{Exp~E (three-stage curriculum) benefits:}} \\
Exp~E $-$ Exp~A (real test)            & $0.9223 - 0.8672$    & $+0.0551$ \\
Exp~E $-$ Exp~B (real gloves)          & $0.8624 - 0.7400$    & $+0.1224$ \\
\bottomrule
\end{tabular}
\vspace{-15pt}
\label{tab:exp-f-gaps}
\end{table}





\section{Conclusion}\label{sec:conclusion}

Full-image metrics (PSNR/SSIM/LPIPS) show the inpainted images are \textit{nearly indistinguishable} from the originals.
Restricting the same metrics to the YOLO-labeled hand crops confirms that the changes are concentrated in the hands as intended.
MediaPipe Hands corroborates this anatomically.
Boundary-gradient analysis shows seam ratios concentrated near $1.09$, indicating visible seams in a very small fraction of images.
Together, the image comparison metrics establish that the inpainting succeeds at the pixel level but introduces a strong, machine-detectable signature.

Four YOLOv8n training schedules (Exps~A, C, D, E), evaluated together with a robustness test (Exp~B) and a cross-experiment analysis (Exp~F) on a paired real/synthetic image dataset, examine whether generative inpainting can close the distribution shift between bare and accessory-bearing hands that public hand-detection datasets present.
The two-stage augment-then-fine-tune schedule (Exp~C) shows that inpainted hand images carry real deployment value.
It gains $8.7$ mAP@0.5 points over the real-only baseline on the standard real test set without sacrificing strict-IoU performance, and closes $18.1$ points of the real-gloves out-of-distribution (OOD) gap.
The three-stage curriculum (Exp~E) narrows progressively from real to real-gloved data and reaches the highest mAP@0.5:0.95 in the study, trading a small amount of mAP@0.5 value for the best box-tightness on both real and real-gloves test splits.

On the $1{,}000$-frame real-gloves test split, Exp~C reaches mAP@0.5 $=0.921$ and Exp~E reaches $0.862$.
However, the strict-IoU component of the real-gloves gap remains persistent: mAP@0.5:0.95 on real gloves is on average $\sim\!0.35$ in these experiments.
This is well below $\sim\!0.62$ that those same models achieve on the standard real test set.
Box tightness on real gloved hands is the genuine residual obstacle and is unlikely to be addressed by further training-procedure changes alone; it likely requires improvements to the synthetic-generation pipeline itself.

\begin{credits}

\subsubsection{\ackname} This study and the acquisition of the dataset were financed by the German Federal Ministry of Labor and Social Affairs (BMAS).
The research was conducted by the junior research group "Artificial Intelligence in a Safe and Healthy Working Environment" at the Federal Institute for Occupational Safety and Health (BAuA).
The authors also thank the Fraunhofer Institute for Manufacturing Engineering and Automation (IPA), Germany, for providing the original (raw) dataset.

\subsubsection{Data and code availability.}
The analysis code and per-experiment results are available at \url{https://github.com/atmikabhardwaj/Schedule_Sensitive_Evaluation_of_Generative_Data}.
The dataset (real and inpainted images with YOLO annotations), trained weights, and training logs are archived at \url{https://doi.org/10.5281/zenodo.21705300}.

\subsubsection{\discintname}
The authors have no competing interests to declare that are relevant to the content of this article.
\end{credits}

\vspace{-8pt}
\bibliographystyle{splncs04}
\bibliography{references}
\end{document}